\documentclass[twoside]{article}

\usepackage{aistats2025}
\usepackage[round]{natbib}

\usepackage[utf8]{inputenc} 
\usepackage[T1]{fontenc}    
\usepackage{hyperref}   
\usepackage{url}            
\usepackage{booktabs}       
\usepackage{amsfonts}       
\usepackage{soul}
\usepackage{amsmath}
\usepackage{amssymb}
\usepackage{amsthm}
\usepackage{nicefrac}       
\usepackage{microtype}      
\usepackage{xcolor}   
\usepackage{graphicx}
\usepackage{algorithm}
\usepackage{float}
\usepackage[normalem]{ulem}
\usepackage{subfigure}
\def\rri{\mathtt{u}}
\def\rris{\rri_{0}}
\def\gmi{\mathtt{b}}
\def\gmiid{\mathtt{g}_{1}}

\def\ex{\mathrm{e}}

\def\xx{\mathtt{x}}

\def\nunu{\nu_{0}}

\def\P{\mathbb{P}}

\newcommand{\wh}{\widehat}

\newtheorem{theorem}{Theorem}[section]
\newtheorem{corollary}{Corollary}[theorem]
\newtheorem{lemma}[theorem]{Lemma}
\DeclareMathOperator{\diag}{diag}

\begin{document}
\pagenumbering{arabic}
%

%

\twocolumn[

\aistatstitle{Causal Discovery on Dependent Binary Data}

\aistatsauthor{ Alex Chen \And Qing Zhou}

\aistatsaddress{Department of Statistics and Data Science \\ University of California, Los Angeles \\ aclheexn1346@g.ucla.edu \And  Department of Statistics and Data Science \\University of California, Los Angeles \\  zhou@stat.ucla.edu} 
]

\begin{abstract}
    The assumption of independence between observations (units) in a dataset is prevalent across various methodologies for learning causal graphical models. However, this assumption often finds itself in conflict with real-world data, posing challenges to accurate structure learning. We propose a decorrelation-based approach for causal graph learning on dependent binary data, where the local conditional distribution is defined by a latent utility model with dependent errors across units. We develop a pairwise maximum likelihood method to estimate the covariance matrix for the dependence among the units. Then, leveraging the estimated covariance matrix, we develop an EM-like iterative algorithm to generate and decorrelate samples of the latent utility variables, which serve as decorrelated data. Any standard causal discovery method can be applied on the decorrelated data to learn the underlying causal graph. We demonstrate that the proposed decorrelation approach significantly improves the accuracy in causal graph learning, through numerical experiments on both synthetic and real-world datasets.
\end{abstract}

\section{Introduction}\label{sec:intro}

Causal discovery methods designed for observational data address the challenges in causal inference under non-experimental settings across many applied domains.
Directed acyclic graphs (DAGs), where nodes correspond to random variables, are a class of graphical models for modeling causal relationships among a set of variables.
Suppose there are $p$ variables and let $PA_j$ be the parent variables of node $j$. Consider a data matrix $X \in \mathbb{R}^{n \times p}$, where each row $x_i\in\mathbb{R}^p$ corresponds to one of the $n$ units in the data. Let $X_j$ be the $j$th column of $X$, which is generated by a structure equation model (SEM) as shown in~\eqref{eq:1} with an independent error vector $\varepsilon_j$: 
\begin{align}
    X_j = f_j(PA_j, \varepsilon_j), \quad \varepsilon_j = (\varepsilon_{1j},...,\varepsilon_{nj}) \overset{\mathrm{iid}}{\sim} \mathbb{P}, \label{eq:1}
\end{align}
for $j \in [p]:=\{1,\dots,p\}$. 

Note that the error terms $\varepsilon_{ij}$, $i\in[n]$ are independent and identically distributed with some distribution $\mathbb{P}$.

A main assumption of this model is the joint independence among the $n$ units of the data matrix $X$ due to the independence of the errors.
More specifically, under the i.i.d assumption, the rows (units) $x_1,\ldots,x_n$ of the data matrix $X$ are independent and thus the $n \times n$ covariance matrix among the $n$ units is diagonal. It is important to distinguish the between-unit covariance matrix from the $p\times p$ covariance matrix among the columns $X_1,\ldots, X_p$, which is typically dense. In this work, the primary motivation is to consider potential dependence among the rows $x_1,\ldots,x_n$ in the learning of the underlying DAG.

Under the i.i.d data assumption, many causal structure estimation methods have been developed. There are three main types of structure learning: constraint-based, score-based, and hybrid methods. One of the primary constraint-based methods is the PC Algorithm \citep{spirtes2000causation}, which uses conditional independence tests to remove edges from an initial complete undirected graph and then orients the edges based on the Meek's rules \citep{10.5555/2074158.2074204}. Variants include PC-select by \citet{buhlmann2010variable} for causal structure estimation on one variable, PC-stable from \citet{colombo2014order} for order-independence, FCI by \citet{spirtes2000causation} for latent confounders, and rankPC by \citet{harris2013pc} for non-normal data. Score-based methods search over a certain graph space to find a graph that optimizes a scoring function such as BIC \citep{BIC} or minimum-description length \citep{Roos2017}. 
Popular examples include the Greedy Equivalence Search (GES) \citep{chickering2002optimal} and the Greedy Hill Climbing \citep{gamez2011learning}. Other variants include Fast Greedy Equivalence Search (FGES) \citep{ramsey2017million}
and regularized likelihood maximization approaches \citep{aragam2015concave,gu2019penalized,fu2013learning}.
Hybrid methods combine the two approaches, constraint and score-based learning, such as MMHC (Max-min Hill climbing) \citep{tsamardinos2006max} and GFCI \citep{ogarrio2016hybrid}. A more complete overview of existing methods for causal learning is provided by  \citet{glymour2019review} and \citet{nogueira2022methods}.

\subsection{Motivation and contributions}\label{sec:motivation}

Real-world data, however, often deviates from the i.i.d assumption. In social and behavioral sciences, the "non-iidness", also termed as \textit{couplings}~\citep{cao2015coupling}, manifests in various forms. Individuals often share characteristics within social groups or families. Consequently, data points become interdependent, breaking the i.i.d assumption. In single-cell RNA sequencing (scRNA-seq), we seek to uncover gene regulatory networks (GRNs) which can be conceptualized as causal networks among genes. Cells in scRNA-seq data have inherent dependence due to various spatial and temporal associations among the cells, such as cell differentiation from the same population. The dependence among units in such data clearly violates the i.i.d. assumptions used in traditional causal discovery methods.

Rather than adapting a specific structure learning method to handle data dependence, we propose a general approach to transform dependent data into an independent surrogate, so that many existing structure learning methods can be applied to the independent surrogate data. More specifically, the main contributions of this work are as follows. First, we propose a DAG model for dependent binary data based on a latent utility model, where errors $\varepsilon_j$ across units are modeled by an unknown covariance matrix $\Sigma$. Then, we develop a pairwise maximum likelihood method to estimate the covariance among the units the data. Lastly, given the estimated $\Sigma$, we develop an EM algorithm to generate surrogate independent data, which are samples of the latent utility variables in our model. We demonstrate, with both synthetic and real-world datasets, that structure learning on the surrogate data is much more accurate than on the original dependent data using state-of-the-art methods.

\subsection{Related works}\label{sec:model_net} 

In the context of DAG-based causal inference, there are only a few very recent methods that take into account the potential
dependence among units.
\citet{pmlr-v115-bhattacharya20a} developed a method for causal inference under \textit{partial interference}~\citep{hudgens2008toward}. They assume a known causal DAG among the variables and then estimate causal effects under dependence among the units $x_i$. In contrast, we do not assume knowledge of a DAG structure and in fact our primary goal is to learn the underlying causal DAG. 
\citet{li2019learning} proposed a linear Gaussian DAG on network data, by introducing dependent exogenous variables $\varepsilon_j = (\varepsilon_{1j},\dots,\varepsilon_{nj})$:
\begin{align}
\begin{split}
    X_j = \sum_{k \in PA_j} \beta_{kj}X_k + \varepsilon_j, \quad \varepsilon_j \sim \mathcal{N}_n (0, \Sigma),
    \label{eq:model}
\end{split} 
\end{align}
where $\beta_{kj}$ is the causal effect of $X_k$ on  $X_j$ and the covariance matrix $\Sigma \in \mathbb{R}^{n \times n}$ is positive definite. They suggest estimating the precision matrix $\Theta=\Sigma^{-1}$ and using the Cholesky factor $L$ of $\Theta$ as a means to decorrelate the data matrix $X$, that is, the decorrelated data $\widetilde{X}=L^{\intercal}X$. This approach demonstrated promising results in addressing data dependence in DAG learning, however, there are a few intrinsic limitations. First, their decorrelation approach is not applicable to discrete data, since $L^{\intercal}X$ is in general outside of the discrete data domain and this transformation has no clear interpretation. Second, \citet{li2019learning} assume that the support of $\Theta$ is restricted to a \textit{known} network (graph) among the $n$ units, which is somewhat limited for many forms of dependent data. 
The wide use and availability of discrete data prompts a novel decorrelation approach. In this work, we introduce a different data-generating mechanism for dependent discrete data and develop an associated decorrelation method for improving causal graph estimation. We do not assume a known network among the units. Our method generates continuous proxy data in the process of removing cross-units dependence, on which a standard structure learning algorithm can be applied to estimate the underlying DAG. As shown by our numerical results, this approach substantially improves the structure learning accuracy and provides a much better model fit to the data.

\section{A DAG model for dependent data}\label{sec:deplatentmodel}



To generalize the SEM in Equation~\eqref{eq:model} to binary variables $x_{ij}\in\{0,1\}$, we use a probit regression model for $[x_{ij}\mid x_{ik},k\in PA_j]$ under a latent-variable formulation.
We introduce a set of auxiliary latent variables  $z_{ij}$ for $j\in[p]$ and $i\in[n]$. 
The binary value of $x_{ij}$ is determined by the latent variable $z_{ij}$:
\begin{align}
    z_{ij} &= \sum_{k\in PA_j}\beta_{kj}x_{ik} + \varepsilon_{ij}=x_i \beta_j +\varepsilon_{ij}, \label{eq:cont} \\ 
    x_{ij} &= I(z_{ij} > 0), \label{eq:discrete}
\end{align}
for $i\in[n]$ and $j\in[p]$, where $\beta_j=(\beta_{1j},\ldots,\beta_{pj})\in\mathbb{R}^p$ such that $\text{supp}(\beta_j)=PA_j$.
Under this formulation, $z_{ij}$ is regarded as the utility for $x_{ij}=1$, while the utility for $x_{ij}=0$ is the baseline (zero). This approach allows us to accommodate dependence among discrete units by assuming
\noindent
\begin{align}\label{eq:cov-dismodel}
\varepsilon_{j} = (\varepsilon_{1j},\dots,\varepsilon_{nj}) \sim \mathcal{N}_n (0, \Sigma), \quad\text{for all }j.
\end{align}
\noindent

Due to an identifiability issue in the model defined by Equations~\eqref{eq:cont} and~\eqref{eq:discrete}, we impose the constraint $\text{diag}(\Sigma)=1$. Without this constraint, the model becomes over-parameterized, and $\Sigma$ and $\beta$ are not identifiable. The dependence between the exogenous variables $\varepsilon_i:=(\varepsilon_{ij}:j\in[p])$ and $\varepsilon_k$ causes the dependence between the two units $x_i$ and $x_k$, $i,k\in[n]$. This assumption aligns with the fact that the exogenous variables are the source of randomness in the general SEM~\eqref{eq:1}. 

Let $Z=(z_{ij})_{n\times p}$ and  $Z_j=(z_{1j},\ldots,z_{nj})$ be the $j$th column. Plugging $x_{ik}=I(z_{ik}>0)$, for all $i=1,\ldots, n$, in Equation~\eqref{eq:cont}, we have
\begin{align}
    Z_j = \sum_{k \in PA_j} \beta_{kj}I(Z_k > 0) + \varepsilon_j, \quad j\in[p]. \label{eq:modelZ}
\end{align}
This defines an SEM for $Z=(Z_1,\ldots,Z_p)$ with a DAG $\mathcal{G}(Z)$.
\begin{lemma}\label{lemma:1}
    Under the latent utility model in Equations~\eqref{eq:cont} and \eqref{eq:discrete}, the DAG $\mathcal{G}(X)$ among the observed discrete variables $X$  is identical to the DAG $\mathcal{G}(Z)$ among the latent variables $Z$.
\end{lemma}
Thus, we develop a method to impute and decorrelate the latent auxiliary data $Z=(Z_{j})$ to work with continuous data rather than discrete.
Then, we apply a standard structure learning method on decorrelated data to learn the underlying DAG. This is the high-level idea of our method.
The key complexity of our model is the interplay between two types of dependence, one over all variables and one over all units.
The causal relations over the variables $X_1,\ldots,X_p$ are modeled through 
a DAG~\eqref{eq:cont} and \eqref{eq:discrete} and the relationship among different units are modeled through a joint
distribution over the exogenous variables $\varepsilon_1,\ldots,\varepsilon_n$~\eqref{eq:cov-dismodel}. 

\textbf{Remark.} Our model does differ from the latent-thresholding model \citep{spirtes1996discovering,silva2005automatic} for discrete variables, where one often assumes a Gaussian DAG for $Z_1,\ldots,Z_p$ and each discrete variable is defined by thresholding, i.e. $X_j=I(Z_j>\tau_j)$. 
In our view, both models are reasonable. Using gene regulatory networks as an example, our model postulates a causal network over the activation/suppression (discrete status) of the genes, while the latent-thresholding model assumes a causal network over the exact expression level (continuous measure) of the genes. When original continuous data is noisy, which is common in many applications, our model could be more robust. 

\section{Methods}\label{sec:methods}
Our main idea is to learn the latent variables $Z_j$. Given the underlying latent data $Z_{j}, j\in[p]$, the causal graph we estimate among $Z$ will be identical to the causal graph among $X$ according to Lemma \ref{lemma:1}. Since $Z$ is continuous, we can apply the Cholesky factor of an estimated precision matrix among the units to remove data dependence.
Our methodology begins by estimating the covariance $\Sigma$ via a pairwise likelihood approach. Given $\widehat \Sigma$, our algorithm iterates between approximating the latent data and estimation of the parameters $\beta_j, j\in[p]$ through an EM-algorithm. The Cholesky factor of the estimated precision matrix can then be used to remove the dependence among the latent data to use for DAG learning by any standard causal structure estimation method. 

\subsection{Covariance estimation} \label{sec:cov_est}
Under classical i.i.d. settings, covariance estimation among discrete variables is a well-studied topic in statistical modeling. \citet{fan2017high} developed a rank based estimator using Kendall's tau to calculate correlations assuming i.i.d samples. \citet{olsson1979maximum} uses a bi-variate normal cdf method to estimate the correlation $\rho_{ij}$ between two discrete variables and also assumes i.i.d samples to obtain an accurate estimate of the correlation. \citet{cui2016copula} uses Gibbs sampling on rank-based data to sample and average correlation matrices from an inverse-Wishart distribution. However, their method does not utilize a prior that can specify sparsity in the covariance matrix. The challenges posed by our model have spurred the necessity to develop a novel covariance estimation approach for our problem, distinguished by several key departures from existing literature. First, our model introduces dependence into discrete data through latent background variables, $\varepsilon_j$, compared to the aforementioned methods where data is assumed to be fully observed. Additionally, since the distribution of $X_j$ depends on its parent variables, we do not have independence among the data involved in the likelihood of $\Sigma$. Furthermore, our method allows for the introduction of sparsity based on potential domain expertise, enhancing practical utility.

Jointly estimating the $n\times n$ covariance matrix $\Sigma$ in Equation~\eqref{eq:cov-dismodel} is very challenging because the error variables $\varepsilon_{ij}$ are unobserved. Therefore, we propose a pairwise MLE method to estimate $\Sigma$. Because $\Sigma_{ii} = 1$ for $i \in [n]$, our pairwise method is applied to estimate each correlation $\rho_{ij}$, for $i,j\in[n]$.

Without loss of generality, let us consider the estimation of $\rho_{12}$, the correlation between the first two units. There are four possible outcomes for $(x_{1j},x_{2j})$ for any $j\in[p]$, i.e. (0,0), (0,1), (1,0), (1,1). Using Equation~\eqref{eq:discrete}, 
\begin{align}
    (x_{1j},x_{2j}) = \left(I\left(\varepsilon_{1j} > -x_1\beta_j \right), I\left(\varepsilon_{2j} > -x_2\beta_j \right)\right). \label{eq:data_gen_trunc}
\end{align}
According to Equation~\eqref{eq:cov-dismodel}, with $\diag(\Sigma) = 1$, the distribution of the two error variables $(\varepsilon_{1j}, \varepsilon_{2j})$, is a bivariate normal with an unknown parameter $\rho_{12}$,
\begin{align}
\begin{pmatrix}\varepsilon_{1j}\\ \label{eq:biv_dist_eq}
\varepsilon_{2j}
\end{pmatrix} &\sim  \mathcal{N}
\begin{bmatrix}
\begin{pmatrix}
0\\
0\\
\end{pmatrix}\!\!,&
\begin{pmatrix}
1 & \rho_{12}\\
\rho_{12} & 1\\
\end{pmatrix}
\end{bmatrix} \hspace{0.5em} \text{for all } j \in [p].
\end{align} 

Given $\beta_j$, we can use 
Equation~\eqref{eq:data_gen_trunc} to find the probability mass function of $(x_{1j}, x_{2j})$ through the CDF of a bivariate Gaussian:
\begin{align*}
    P(x_{1j}, x_{2j}|D_j^{1,2}, \rho_{12}) = \iint\limits_{D_j^{1,2}} \phi \left(u_1,u_2|\rho_{12} \right)du_1 du_2,
\end{align*}
where $D_j^{1,2} \subset \mathbb{R}^2$ is the domain for the integral
and $\phi$ is the pdf for the bivariate normal in Equation~\eqref{eq:biv_dist_eq}. For example, if $(x_{1j}, x_{2j}) = (1,1)$ then the domain $D_j^{1,2}= (-x_{2}\beta_j, \infty) \times (-x_{2}\beta_j, \infty)$, where $\text{supp}(\beta_j)=PA_j$. 
The likelihood of $\rho_{12}$ given all $p$ pairs $(x_{1j}, x_{2j})$, $j\in[p]$ is 
\begin{align}\label{alg:likelihood_cov_mat}
    L(\rho_{12}|x_1,x_2) = \prod_{j = 1}^p P(x_{1j}, x_{2j}|D_j^{1,2}, \rho_{12}). 
\end{align}
For any pair of units $(a,b)\in [n]\times [n]$, our estimate $\widehat\rho_{ab}$ is the maximizer of the likelihood function $L(\rho_{ab}|x_a,x_b)$, which can be found easily using a univariate numerical optimization method. 
 Figure~\ref{fig:rho_likelihood_sim} in Supplementary Material~\ref{supp:A1} illustrates the accuracy of our estimate compared to the true correlation.

The covariance estimation method 
relies on an initial estimate of $\beta$, necessitating initial parent estimates $\widehat{PA}_j$ for $j \in [p]$. 
We apply the Max-Min Hill Climbing (MMHC) algorithm \citep{tsamardinos2006max} to estimate a DAG and then use logistic regression (See Supplementary Material~\ref{supp:A3}) to estimate $\beta_j$ from $PA_j$. Our covariance estimation is robust to the initial estimates, as suggested by the numerical results in Supplementary Material~\ref{supp:A3}.

For practical results, we restrict our attention to covariance matrices with a block-diagonal structure. Block structure often occurs in data with clusters of dependent units. We do not need to assume any sparse structure within each block. We apply the pairwise MLE to each $\rho_{ij}$ between two units in the same block and apply a regularization step to ensure the positive definiteness of $\wh{\Sigma}$. We obtain a positive semi-definite matrix by truncating the negative eigenvalues $\lambda_i$ of our estimated covariance matrix, i.e. $\lambda'_i = \text{max}\{0, \lambda_i\}$ and then re-scale the off-diagonal elements by a factor $<1$ (e.g. 0.9).

\subsection{Latent Data Recovery and Decorrelation}\label{sec:disc_decor_alg}


Given the estimated covariance matrix $\widehat\Sigma$, we develop an  EM algorithm \citep{dempster1977maximum} to recover the latent data and update parameter estimates. In the E-step, given $\beta$, we impute the latent data, $Z$, utilizing draws of $\varepsilon$ from a Truncated-Normal distribution. 
In the M-step, decorrelation is applied to the latent data when maximizing the complete data log-likelihood to update $\beta$. 

With the estimated covariance matrix $\wh{\Sigma}$ from Section~\ref{sec:cov_est}, the distribution of the error term $\varepsilon_j$ is $\mathcal{N}_n(0, \wh\Sigma)$. 
However, the data vector $X_j$ is determined by the relationship between $\varepsilon_j$ and $-X\beta_j$ as specified in Equation~\eqref{eq:data_gen_trunc}. Thus, given $X$ and the current parameter $\beta$, $\epsilon_j$ is distributed 
\begin{align}\label{eq:condeps}
\varepsilon_j\mid X,\beta_j \sim \mathcal{N}_T(0, \wh\Sigma),
\end{align}
which is truncated at $-X\beta_j$. More specifically, the truncation at $-X\beta_j$ introduces a total of $n$ constraints in the truncated normal distribution in the form of
\begin{align*}
\begin{split}
    \varepsilon_{ij} &> -x_i\beta_j  \quad\text{if } x_{ij}=1 \\
    \varepsilon_{ij} &\leq -x_i\beta_j  \quad\text{if } x_{ij}=0.
\end{split}
\end{align*}
We leverage the block structure of the covariance matrix to draw $\varepsilon_j$ by simulating a multivariate truncated normal distribution within each block using a Gibbs sampler. 
The average of the $N$ draws yields an approximate expectation $\mathbb{E}(\varepsilon_j\mid X,\beta_j)$. We can then reconstruct latent auxiliary data $Z$ via Equation~\eqref{eq:cont}.

After reconstructing latent data $Z_j$, $j\in[p]$, we develop a decorrelation method using the estimated covariance matrix $\wh\Sigma$ in the M-step to update each $\beta_j$.
Let $L^{\intercal}$ be the Cholesky factor of $\Theta:=\Sigma^{-1}$ such that $\Theta=L L^{\intercal}$.
Given $\varepsilon_j \sim \mathcal{N}_n(0, \Sigma)$, applying the Cholesky factor $L^{\intercal}$ results in $L^{\intercal}\varepsilon_j \sim \mathcal{N}_n(0,I_n)$, a vector of independent Gaussian variables. Accordingly, for each $j\in[p]$, decorrelation of the latent data $Z_j$ can be performed by applying $\wh L^{\intercal}$ to $Z_j$
such that
\begin{align}\label{eq:decorr}
    \wh{L}^{\intercal}Z_j &= \wh{L}^{\intercal}X\beta_j + \wh{L}^{\intercal}\varepsilon_j,
\end{align}
where $\wh L^{\intercal}$ is the Cholesky factor of $\wh\Sigma^{-1}$.
This simplifies to a regression of $\wh{L}^{\intercal}Z_j$ onto $\wh{L}^{\intercal}X$ to estimate $\beta_j$, but restricting the support of $\beta_j$ to $\wh{PA}_j$. This process maximizes the expectation of the complete data log-likelihood $\log p(X,Z\mid\beta)$ to update $\beta_j$. Ridge regression is used in order to avoid overfit to potentially inaccurate draws of $\varepsilon$.

The iterative process outlined in Algorithm~\ref{alg:discdag} consists of recovering the latent data $Z$, decorrelating $Z$, and estimating $\beta$. 
In Supplementary Material~\ref{supp:A2}, Figure~\ref{fig:converge}, a single example run of this algorithm is shown. Notably, the $\beta$ values begin to converge to a stable point after approximately 5 iterations.
\begin{algorithm}
\caption{EM-based decorrelation algorithm}\label{alg:discdag}
Given an estimated Cholesky factor $\wh L$ from $\wh \Sigma$ and initial estimated graph $\wh{\mathcal{G}}$, iterate between the following steps until a stop criterion is met:
\begin{enumerate}
    \item Average draws of $\wh\varepsilon$ according to Equation~\eqref{eq:condeps}.
    \item Reconstruct latent variable data $\wh{Z} = X\wh{\beta} + \wh{\varepsilon}$.
\item Decorrelate latent data $\wh{L}^{\intercal}\wh{Z} = \wh{L}^{\intercal}X\wh{\beta} + \wh{L}\wh{\varepsilon}$.
\item Perform $p$ ridge regressions of $\wh{L}^{\intercal}\wh{Z}_j$ on $\wh{L}^{\intercal}X_j$ for each $j \in [p]$ to obtain $\wh{\beta}$ where the support of $X_j$ is the parents of variable $j$.
\end{enumerate}
\end{algorithm}
\subsection{Structure Learning} \label{sec:struc_learning}
Lemma \ref{lemma:1} suggests that we may use the imputed $Z$ for causal discovery of $X$. Algorithm~\ref{alg:discdag} produces decorrelated data, $\wh{L}^{\intercal}Z$, at each iteration, where unit-dependence has been largely removed. With close-to-independent data, conventional structure learning methods can be employed to estimate a DAG among $Z$. 
Due to Monte Carlo simulation of $\varepsilon$, the $\wh Z$ is not an exact expectation. We mitigate this inaccuracy by
using a consensus or average across multiple imputations of $Z$ in DAG learning.

Since we are using observational data, in general, one can only learn an equivalence class represented by a CPDAG. We employ two approaches to generate the final estimated CPDAG. In the first approach, we run a standard structure learning algorithm with default parameters on $M=10$ decorrelated datasets imputed at different iterations of Algorithm~\ref{alg:discdag}. We accept an edge into our final graph estimate if at least half of the estimated CPDAGs agree. We call this the \textit{consensus approach}. In the second approach, we average all $M$ decorrelated datasets and then run a standard structure learning algorithm on the averaged dataset, which we call the \textit{average approach}. From a computation time perspective, \textit{average approach} necessitates only a single DAG structure estimate while the \textit{consensus approach} requires $M$ estimates. We compare these two approaches to a \textit{baseline approach} that uses a standard structure learning algorithm directly to the discrete dependent data.

\section{Experimental Results} \label{sec:data_sim}

For simulations, we use either random DAGs or those from the \textit{bnlearn} repository. Given a DAG structure, edge weights were uniformly sampled from the interval $[-0.9,-0.6] \cup [0.6,0.9]$. We then sampled noise variables, $\varepsilon$, according to Equation~\eqref{eq:cov-dismodel} and a specified covariance structure detailed below. We simulate the continuous auxiliary variables $Z_j$ and discretize $Z_j$ using Equation~\eqref{eq:discrete} to obtain our observed data $X_j$. The process is repeated according to a topological ordering until we have discrete data $X_j$ for all $j\in[p]$.

Three covariance structures (of $\Sigma$) among the $n$ units were employed based on a block-diagonal structure. Within each cluster (block), the units were correlated following one of the structures where block sizes range from 10 to 15 in simulated data: (i) \textit{Equal} covariance implies fully-connected units in a cluster with $\Sigma_{ij} = \theta$ if $i \neq j$ where $\theta \sim \mathcal{U}(0.4, 0.7)$, (ii) \textit{Toeplitz} covariance means units are connected in a Markov Chain, with $\Sigma_{ij} = \theta^{|i - j|/5}$ where $\theta \sim \mathcal{U}(0.1,0.25)$, and (iii) \textit{Mixed} covariance means each block is randomly \textit{Toeplitz} or \textit{Equal}. To validate the covariance estimation accuracy, we constructed 10 random DAGs for each pair of parameters $n \in\{100, 500\}$ units and $p \in \{100, 1000\}$ variables and simulated data according to aforementioned data generating process using a mixed covariance structure for $\Sigma$. 
After obtaining the initial estimates of $\wh{\beta}$, we apply the pairwise MLE to estimate each $\rho_{ij}$ within the known block structure.  
Then, we calculated the RMSE of $\wh{\Sigma}$ with respect to the true covariance matrix $\Sigma^*$ for each random DAG. As reported in Supplementary Material~\ref{supp:A6}, our covariance estimates have low RMSE values $\leq 0.15$ for $p = 100$ and $\leq 0.06$ for $p = 1000$. 

Using the estimated covariance matrix $\wh{\Sigma}$, Algorithm~\ref{alg:discdag} is used to remove the dependence from the data. As mentioned in Section~\ref{sec:struc_learning}, we took the approximately decorrelated datasets and used two approaches, \textit{consensus} and \textit{average}. To ensure the results of our algorithm were consistent across different structure learning methods, we chose three different methods, MMHC~\citep{tsamardinos2006max}, PC, and Copula-PC~\citep{cui2016copula}. Copula-PC is an extension of the traditional PC algorithm for
non-Gaussian data using copulas. Specifically, we use the discrete versions of each learning method on the dependent discrete data and the continuous versions on the decorrelated data. Then, we compared the estimated CPDAG against the underlying true CPDAG using an F-1 score.

\begin{figure*}[t]
    \centering
    \includegraphics[width=0.95\linewidth]{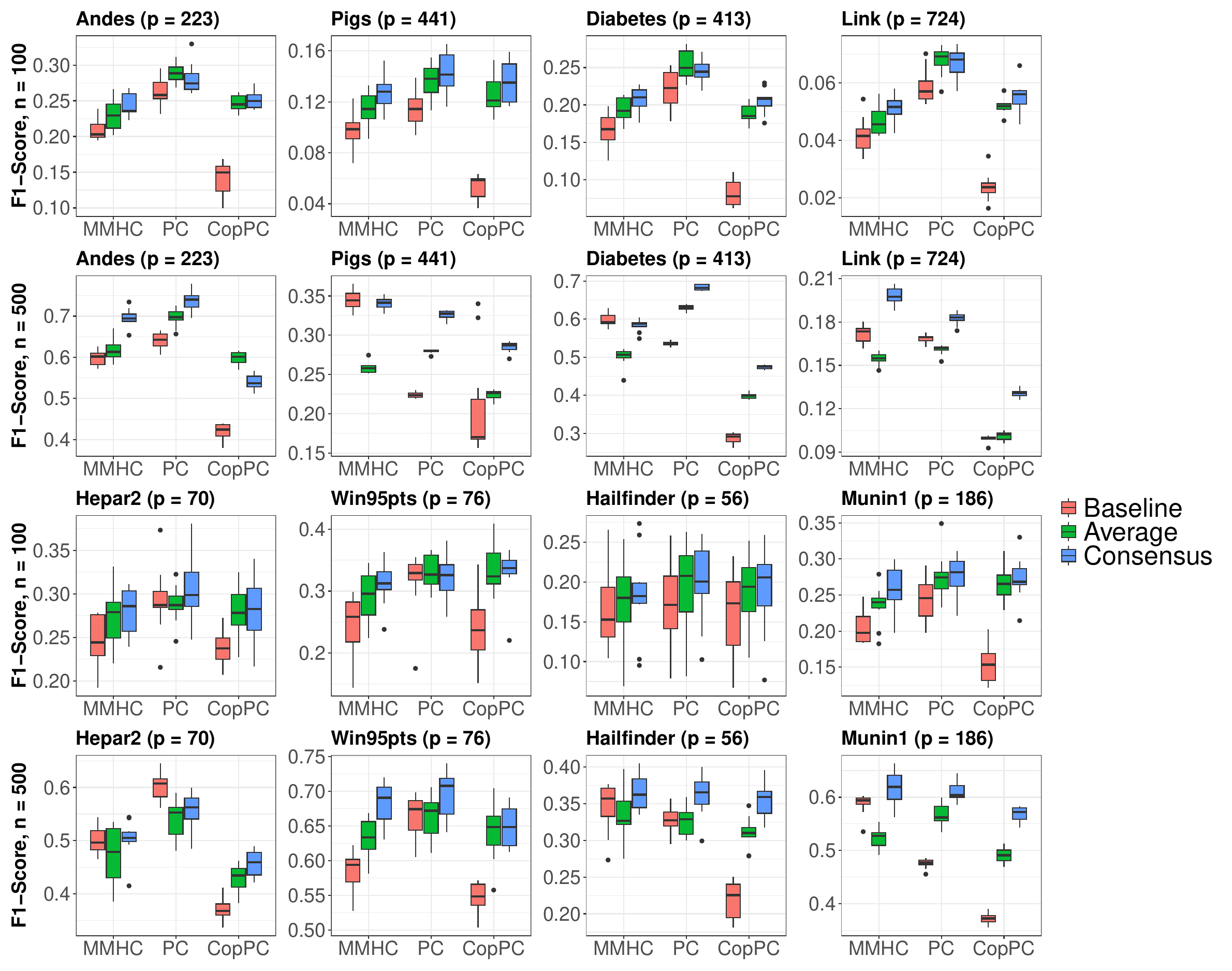}\hspace*{-2cm}
\caption{Structure learning accuracy before and after decorrelation for 8 real Bayesian networks.}
\label{fig:bn_results}
\end{figure*}

Eight real DAGs ($p$ variables and $s$ edges) were sourced from the bnlearn repository \citep{scutari2014bayesian}: \textbf{Hailfinder} ($p = 56, s = 66$), \textbf{Hepar2} ($p = 70, s = 123$), \textbf{Win95pts} ($p = 76, s = 112$), \textbf{Munin1} ($p = 186, s = 273$), \textbf{Andes} ($p = 223, s = 338$), \textbf{Pigs} ($p = 441, s = 592$), \textbf{Diabetes} ($p = 413, s = 602$), \textbf{Link} ($p = 724, s = 1125$). These DAGs cover various domains, such as severe hail forecasting in northeastern Colorado from \textbf{Hailfinder} and the use of an electromyography machine in medical diagnostics in \textbf{Munin}. We used the aforementioned parameters for each simulation: $n \in \{100, 500\}$ units, edge weights $\beta$ were uniformly sampled from $[-0.9,-0.6] \cup [0.6,0.9]$ and the covariance matrix $\Sigma^*$ was sampled according to the mixed structure with block sizes ranging from 10 to 15. With these parameters, we simulated data for each DAG, repeating the process 10 times.

Figure~\ref{fig:bn_results} reports an F-1 score of estimated CPDAGs to the true CPDAG after using each approach. Red boxplots serve as the \textit{baseline} and the green and blue boxplots correspond to the \textit{average} and \textit{consensus} approaches, respectively. Our Consensus method demonstrates on average 20\% improvement to the F-1 score compared to the baseline method, with only three simulations displaying exceptions.

Our real DAG results concur with additional experiments under the same settings as above for random DAGs with $n \in \{100, 500\}$ units, $p \in \{100,1000\}$ variables and $2p$ edges. Figure~\ref{fig:simulated_hist_f1} in Supplementary Material~\ref{supp:randomDags} demonstrates between 13-34\% improvement for random DAGs with either the \textit{consensus} or \textit{average} approach over the \textit{baseline} method. For \textit{Toeplitz} and \textit{Equal} covariance structures and a more detailed numerical comparison, see Supplementary Material~\ref{supp:A3}.

\begin{figure}
    \centering
    \includegraphics[width=1\linewidth]{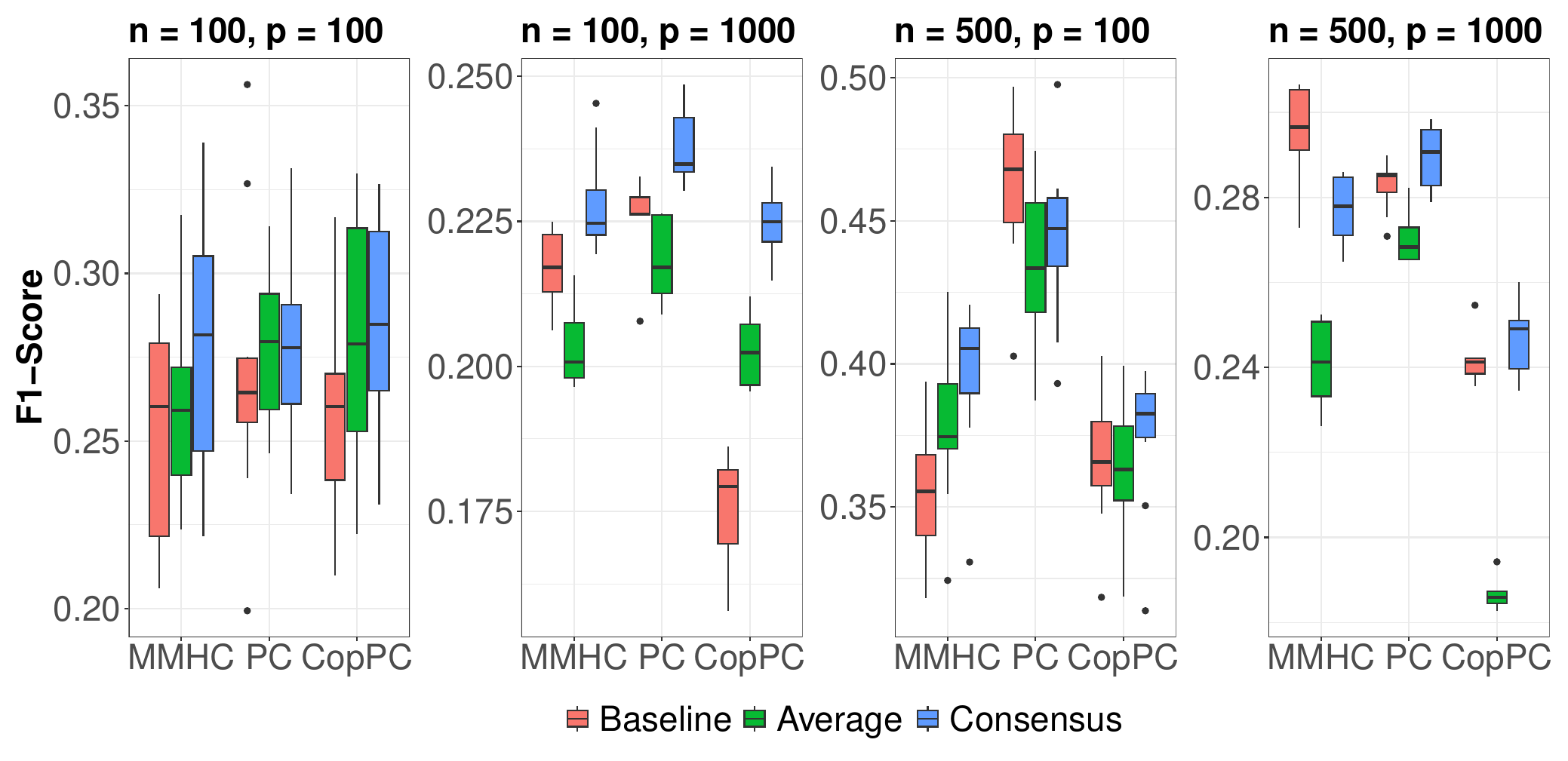}
    \caption{F-1 scores before and after decorrelation across 10 simulations for each setting of $(n,p)$ under deviations from our model assumptions.}
    \label{fig:nl_sim}
\end{figure}

In practical application, robustness to violations of model assumptions is very important. Thus, we performed additional simulations where our model assumptions are violated. We assume $Z_1,\ldots,Z_p$ follow a nonlinear DAG model: 
\[z_{ij}=\sum_{k\in PA_j}f_{kj}(z_{ik}) + \varepsilon_{ij},\] 
where, for each edge $k\to j$, $f_{kj} = (\beta_{kj} z_{ik})^2$ with probability $\frac{1}{2}$ and $f_{kj}(z_{ik}) = \beta_{kj} z_{ik}$ with probability $\frac{1}{2}$. The error vector $\varepsilon_j$ follows the same $n$-variate joint Gaussian in Equation~\eqref{eq:cov-dismodel}. Then, the observed variables are $x_{ij}=I(z_{ij}>\tau_j)$, where $\tau_j$ is a cutoff chosen as the median of $Z_j$. There are two deviations from the model assumptions specified in Equations~\eqref{eq:discrete} and \eqref{eq:cov-dismodel}. First, the parent set of $Z_j$ are $X_{PA_j}$ in Equation~\eqref{eq:discrete}, compared to $Z_j$'s parents being $Z_{PA_j}$ in this simulation. Second, we assume nonlinear relations in this simulation. As shown in Figure~\ref{fig:nl_sim}, for most cases the consensus approach was able to improve the accuracy of the three DAG learning methods. This demonstrates that our decorrelation method is quite robust to violations against the model assumptions.

\section{Application on scRNA-Seq Data}

With recent advancements in single-cell RNA sequencing technology, researchers can measure gene expression for thousands of cells. 
We seek to learn  gene regulatory networks (GRNs) that encode the causal relationships governing gene expression in different biological processes. Each node in a GRN represents a gene and a direct edge from gene $X$ to $Y$ indicates a direct regulatory effect of $X$ on $Y$.
In this study, we use an RNA-seq dataset published in \citet{chu2016single}, 
which consists of gene expression measurements from approximately 20,000 genes across $n = 1018$ cells. The cells in this data include undifferentiated human embryonic stem cells (hESCc), different lineage-specific progenitor cells that can differentiate into specific cell types,
and fibroblasts that served as a control sample. Since lineage-specific progenitors were differentiated from the same population of hESCs, dependence among these cells is highly expected.
We processed the data following the methodology outlined by \citet{li2018modeling} involving imputation of missing values and application of a log transformation. 
Our study focuses on the estimation of a GRN among a subset of $p = 51$ target genes selected by \citet{chu2016single}. The remaining genes are  referred to as background genes hereafter.
\subsection{Pre-estimate of Block Structure}\label{sec:scrna_preest}

The distinct cell types in the data, by experimental design, suggest that a block structure for the covariance matrix $\Sigma$ among the cells.
However, cells of the same type may not necessarily belong to the same block. Therefore, we applied hierarchical clustering with complete linkage, utilizing 2,000 background genes as the feature vector to partition the cells into clusters. The resulting dendrogram was cut to ensure that a majority of clusters were comprised of at least $15$ cells. From each cluster, we randomly sampled 15 to 30 cells and subsequently defined the block structure for $\Sigma$ by these clusters. 
Distinguishing between activated and suppressed genes may be a more robust representation than using continuous or count data. Therefore, we use k-means to identify a natural threshold for discretizing the data based on the distribution of expression counts rather than imposing an arbitrary threshold, transforming the data into binary states (0 or 1). After pre-processing, our dataset consists of $n = 384$ cells, distributed across 14 blocks of 15 to 30 cells, with expression measure for the $p = 51$ target genes.  

\subsection{Model Evaluation} \label{sec:modeleval}

As the true underlying GRN is unknown, direct evaluation of our estimates is unattainable. Instead, we assess each method by evaluating the test data likelihood of an estimated graph through cross-validation. To make use of the independence between cell blocks, each CV fold aligned with an estimated block. 
The primary challenge lies in estimating the covariance matrix $\Sigma$ for the test data. Using test data would introduce bias in the likelihood evaluation. To address this, we randomly sampled 100 background genes and ran the covariance estimation method using these genes to obtain a pre-estimated $\widetilde\Sigma$, which was then used to evaluate the likelihood of each test dataset.
Note that no part of the test data was involved in the estimation of $\widetilde\Sigma$. 

\begin{figure}
    \centering
    \includegraphics[width=1\linewidth]{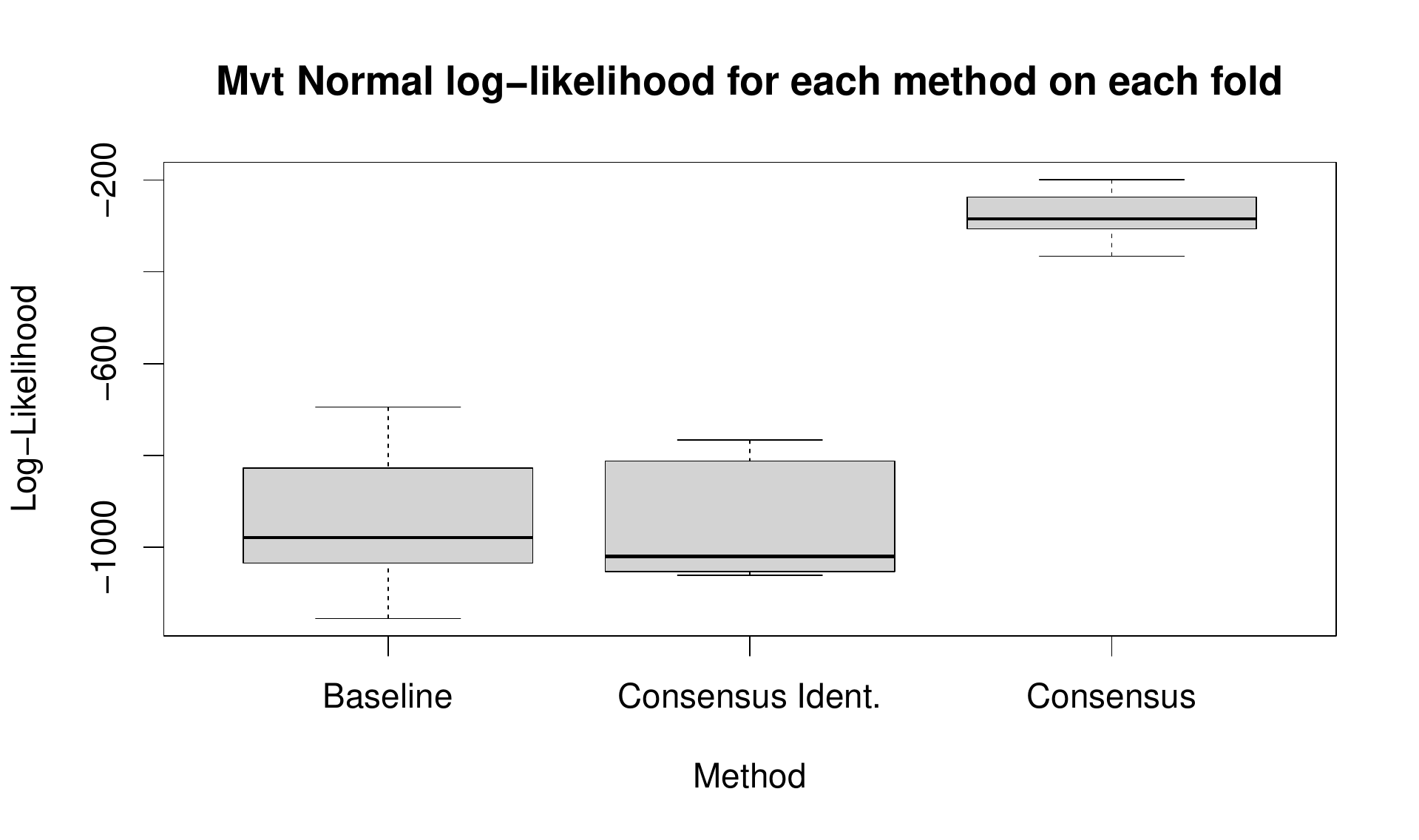}
    \caption{Distribution of the test-data log-likelihood across the 10 folds in the RNA-seq data using three different methods.}
    \label{fig:Mvt_likelihood_realdata}
\end{figure}

We compare three approaches on this dataset. Each approach estimates a graph $\wh{\mathcal{G}}$ and the associated parameters from the training data and then evaluates the likelihood of the test data. This process is repeated across 10 folds of CV.
The \textit{baseline} method estimates $\wh{\mathcal{G}}$ using the MMHC method on the discrete data. 
The second method, \textit{consensus ident.} uses Algorithm~\ref{alg:discdag} with $\wh L=I$, essentially assuming no dependence between the cells. For the third method, we decorrelate latent data and apply the \textit{consensus} approach to estimate $\wh{\mathcal{G}}$. 
As shown in Figure~\ref{fig:Mvt_likelihood_realdata}, the \textit{consensus} approach which considers cell dependence best fits the test data across all CV folds, with a substantial margin from the other two methods. We calculate a normalized likelihood ratio as a comparative tool, which can be defined as the average likelihood ratio of observing a single data point. This is expressed as
        $\left\{{P(D | \text{Model}_m)}/{P(D | \text{Model}_b)}\right\}^{1/np},$
where $D$ represents the test data, $\text{Model}_m$ is the model estimated from our method, and $\text{Model}_b$ is the baseline estimated model. Calculating this metric using the median test data log-likelihoods in Figure~\ref{fig:Mvt_likelihood_realdata} results in a ratio of $1.57$ between the \textit{consensus} method and the \textit{baseline} method and a ratio of $1.62$ between the \textit{consensus} method and the \textit{consensus ident.} method.
Experiments with varying cluster sizes and background genes yielded consistent results. The only difference between \textit{consensus ident.} and \textit{consensus} in Figure~\ref{fig:Mvt_likelihood_realdata} is the use of the pre-estimated covariance matrix for decorrelation, emphasizing the strong impact of  between-cell dependence on fitting a graphical model. This demonstrates that our proposed dependent DAG model fits this RNA-seq dataset much better and confirms the significance of capturing dependence among cells. It shows the dependence
can be well-estimated from a random set of genes, supporting our assumption of using the same $\Sigma$ for all background variables $\varepsilon_j$, $j\in [p]$.

\section{Discussion} 
In this work, we developed the idea of data decorrelation for DAG learning on binary data. The key components include a pairwise MLE for covariance estimation and an iterative algorithm for generating and decorrelating surrogate continuous data. 

\subsection{Summary} \label{sec:conclusions}
Independence in data is commonly assumed in practical application where it does not hold. Extensive experiments on both synthetic and real data 
 using our method for dependent data showcase significant improvements over existing methods, particularly in cases with $p > n$ and in real RNA-seq data for GRN estimation. We believe our algorithm for causal estimation of dependent units is particularly easy for practical application as users do not need to set additional parameters other than parameters necessary for classical causal discovery methods such as PC or MMHC. The improvement in estimation of causal models is important for applications such as GRNs, providing unexplored avenues for experimentation in drug discovery.

Our experiments reveal that our decorrelation methodology exhibits superior performance in scenarios with a strong underlying correlation structure, while demonstrating comparable performance in other settings. We advocate for using the proposed method in instances where there exists a discernible dependence structure within the dataset or when the sample size is not too large (Due to computation size). To determine the presence of a strong correlation, one practical approach could involve testing if the correlation coefficient $\rho_{ab}$ between rows $a$ and $b$ is significantly different from 0.

\subsection{Limitations and future work} \label{sec:limits}
We acknowledge a few limitations in our algorithm and experiments. First, our simulations are limited to binary data, which restricts the applicability of our method to real-world data scenarios where multiple categories or mixed data can be more common. Thus, considering a mix of continuous and discrete data could increase the applicability of our method to real data.

Our method can be modified to accommodate multi-category discrete and mixed data settings. Generalizing to multiple categories is feasible but introduces a few computational challenges. Suppose $x_{ij}\in\{0,\ldots,K\}$. We introduce a set of cutoff values, $\tau_{jk}$, $k=0,\ldots,K+1$ where $\tau_{j0}=-\infty$ and $\tau_{j,K+1}=\infty$. 
Let $x_{ij}=k$ if and only if $\tau_{jk}\leq z_{ij}<\tau_{j,k+1}$ in place of Equation~\eqref{eq:discrete}. Then, we can generalize the covariance estimation method in Section~\ref{sec:cov_est} by including the cutoffs $\{\tau_{jk}\}$ as unknown parameters. Once $\Sigma$ and $\{\tau_{jk}\}$ have been estimated, the same decorrelation approach can be applied to simulate the latent continuous variables $z_{ij}$ as described in Section~\ref{sec:disc_decor_alg}.  The key computational challenges lie in the covariance estimation step, where the number of boundaries/regions increases quadratically with each additional category, complicating the estimation process. Similarly, simulating $\varepsilon_j$ from a multivariate truncated normal becomes more complex due to the high-dimensional space constrained by an increasing number of boundaries. For mixed data, if $X_j$ is continuous, we adjust Equation~\eqref{eq:discrete} by $x_{ij}=z_{ij}$. Then, we modify the likelihood in Equation~\eqref{alg:likelihood_cov_mat}, where $(x_{1j},x_{2j})$ follows a bivariate normal without truncation if they are continuous. Using a similar pairwise MLE we can estimate the covariance matrix $\Sigma$, whose Cholesky factor $\widehat{L}^T$ can be used to decorrelate the data. For continuous $X_j$, decorrelation is achieved simply by $\widehat{L}^T X_j$, while the decorrelated data for binary variables is given by $\widehat{L}^T Z_j$ in Equation~\eqref{eq:decorr}. Then we may apply a standard DAG learning method to the decorrelated data where all columns are continuous. 

\newpage
\bibliographystyle{plainnat}
\bibliography{aistats-2025}

\begin{thebibliography}{31}
\providecommand{\natexlab}[1]{#1}
\providecommand{\url}[1]{\texttt{#1}}
\expandafter\ifx\csname urlstyle\endcsname\relax
  \providecommand{\doi}[1]{doi: #1}\else
  \providecommand{\doi}{doi: \begingroup \urlstyle{rm}\Url}\fi

\bibitem[Aragam and Zhou(2015)]{aragam2015concave}
Bryon Aragam and Qing Zhou.
\newblock Concave penalized estimation of sparse gaussian bayesian networks.
\newblock \emph{The Journal of Machine Learning Research}, 16\penalty0 (1):\penalty0 2273--2328, 2015.

\bibitem[Bhattacharya et~al.(2020)Bhattacharya, Malinsky, and Shpitser]{pmlr-v115-bhattacharya20a}
Rohit Bhattacharya, Daniel Malinsky, and Ilya Shpitser.
\newblock Causal inference under interference and network uncertainty.
\newblock In Ryan~P. Adams and Vibhav Gogate, editors, \emph{Proceedings of The 35th Uncertainty in Artificial Intelligence Conference}, volume 115 of \emph{Proceedings of Machine Learning Research}, pages 1028--1038. PMLR, 22--25 Jul 2020.
\newblock URL \url{https://proceedings.mlr.press/v115/bhattacharya20a.html}.

\bibitem[B{\"u}hlmann et~al.(2010)B{\"u}hlmann, Kalisch, and Maathuis]{buhlmann2010variable}
Peter B{\"u}hlmann, Markus Kalisch, and Marloes~H Maathuis.
\newblock Variable selection in high-dimensional linear models: partially faithful distributions and the pc-simple algorithm.
\newblock \emph{Biometrika}, 97\penalty0 (2):\penalty0 261--278, 2010.

\bibitem[Cao(2015)]{cao2015coupling}
Longbing Cao.
\newblock Coupling learning of complex interactions.
\newblock \emph{Information Processing \& Management}, 51\penalty0 (2):\penalty0 167--186, 2015.

\bibitem[Chickering(2002)]{chickering2002optimal}
David~Maxwell Chickering.
\newblock Optimal structure identification with greedy search.
\newblock \emph{Journal of machine learning research}, 3\penalty0 (Nov):\penalty0 507--554, 2002.

\bibitem[Chu et~al.(2016)Chu, Leng, Zhang, Hou, Mamott, Vereide, Choi, Kendziorski, Stewart, and Thomson]{chu2016single}
Li-Fang Chu, Ning Leng, Jue Zhang, Zhonggang Hou, Daniel Mamott, David~T Vereide, Jeea Choi, Christina Kendziorski, Ron Stewart, and James~A Thomson.
\newblock Single-cell rna-seq reveals novel regulators of human embryonic stem cell differentiation to definitive endoderm.
\newblock \emph{Genome biology}, 17:\penalty0 1--20, 2016.

\bibitem[Colombo et~al.(2014)Colombo, Maathuis, et~al.]{colombo2014order}
Diego Colombo, Marloes~H Maathuis, et~al.
\newblock Order-independent constraint-based causal structure learning.
\newblock \emph{J. Mach. Learn. Res.}, 15\penalty0 (1):\penalty0 3741--3782, 2014.

\bibitem[Cui et~al.(2016)Cui, Groot, and Heskes]{cui2016copula}
Ruifei Cui, Perry Groot, and Tom Heskes.
\newblock Copula pc algorithm for causal discovery from mixed data.
\newblock In \emph{Machine Learning and Knowledge Discovery in Databases: European Conference, ECML PKDD 2016, Riva del Garda, Italy, September 19-23, 2016, Proceedings, Part II 16}, pages 377--392. Springer, 2016.

\bibitem[Dempster et~al.(1977)Dempster, Laird, and Rubin]{dempster1977maximum}
Arthur~P Dempster, Nan~M Laird, and Donald~B Rubin.
\newblock Maximum likelihood from incomplete data via the em algorithm.
\newblock \emph{Journal of the royal statistical society: series B (methodological)}, 39\penalty0 (1):\penalty0 1--22, 1977.

\bibitem[E.~Schwarz(1978)]{BIC}
Gideon E.~Schwarz.
\newblock Estimating the dimension of a model.
\newblock \emph{The Annals of Statistics}, 6, 03 1978.
\newblock \doi{10.1214/aos/1176344136}.

\bibitem[Fan et~al.(2017)Fan, Liu, Ning, and Zou]{fan2017high}
Jianqing Fan, Han Liu, Yang Ning, and Hui Zou.
\newblock High dimensional semiparametric latent graphical model for mixed data.
\newblock \emph{Journal of the Royal Statistical Society Series B: Statistical Methodology}, 79\penalty0 (2):\penalty0 405--421, 2017.

\bibitem[Fu and Zhou(2013)]{fu2013learning}
Fei Fu and Qing Zhou.
\newblock Learning sparse causal gaussian networks with experimental intervention: regularization and coordinate descent.
\newblock \emph{Journal of the American Statistical Association}, 108\penalty0 (501):\penalty0 288--300, 2013.

\bibitem[G{\'a}mez et~al.(2011)G{\'a}mez, Mateo, and Puerta]{gamez2011learning}
Jos{\'e}~A G{\'a}mez, Juan~L Mateo, and Jos{\'e}~M Puerta.
\newblock Learning bayesian networks by hill climbing: efficient methods based on progressive restriction of the neighborhood.
\newblock \emph{Data Mining and Knowledge Discovery}, 22:\penalty0 106--148, 2011.

\bibitem[Glymour et~al.(2019)Glymour, Zhang, and Spirtes]{glymour2019review}
Clark Glymour, Kun Zhang, and Peter Spirtes.
\newblock Review of causal discovery methods based on graphical models.
\newblock \emph{Frontiers in genetics}, 10:\penalty0 524, 2019.

\bibitem[Gu et~al.(2019)Gu, Fu, and Zhou]{gu2019penalized}
Jiaying Gu, Fei Fu, and Qing Zhou.
\newblock Penalized estimation of directed acyclic graphs from discrete data.
\newblock \emph{Statistics and Computing}, 29:\penalty0 161--176, 2019.

\bibitem[Harris and Drton(2013)]{harris2013pc}
Naftali Harris and Mathias Drton.
\newblock Pc algorithm for nonparanormal graphical models.
\newblock \emph{Journal of Machine Learning Research}, 14\penalty0 (11), 2013.

\bibitem[Hudgens and Halloran(2008)]{hudgens2008toward}
Michael~G Hudgens and M~Elizabeth Halloran.
\newblock Toward causal inference with interference.
\newblock \emph{Journal of the American Statistical Association}, 103\penalty0 (482):\penalty0 832--842, 2008.

\bibitem[Li et~al.(2024)Li, Padilla, and Zhou]{li2019learning}
Hangjian Li, Oscar Hernan~Madrid Padilla, and Qing Zhou.
\newblock {Learning Gaussian DAGs from Network Data}.
\newblock \emph{Journal of Machine Learning Research, accepted}, 2024.

\bibitem[Li and Li(2018)]{li2018modeling}
Wei~Vivian Li and Jingyi~Jessica Li.
\newblock Modeling and analysis of rna-seq data: a review from a statistical perspective.
\newblock \emph{Quantitative Biology}, 6:\penalty0 195--209, 2018.

\bibitem[Meek(1995)]{10.5555/2074158.2074204}
Christopher Meek.
\newblock Causal inference and causal explanation with background knowledge.
\newblock In \emph{Proceedings of the Eleventh Conference on Uncertainty in Artificial Intelligence}, UAI'95, page 403–410, San Francisco, CA, USA, 1995. Morgan Kaufmann Publishers Inc.
\newblock ISBN 1558603859.

\bibitem[Meinshausen and B{\"u}hlmann(2006)]{meinshausen2006high}
Nicolai Meinshausen and Peter B{\"u}hlmann.
\newblock High-dimensional graphs and variable selection with the lasso.
\newblock 2006.

\bibitem[Nogueira et~al.(2022)Nogueira, Pugnana, Ruggieri, Pedreschi, and Gama]{nogueira2022methods}
Ana~Rita Nogueira, Andrea Pugnana, Salvatore Ruggieri, Dino Pedreschi, and Jo{\~a}o Gama.
\newblock Methods and tools for causal discovery and causal inference.
\newblock \emph{Wiley interdisciplinary reviews: data mining and knowledge discovery}, 12\penalty0 (2):\penalty0 e1449, 2022.

\bibitem[Ogarrio et~al.(2016)Ogarrio, Spirtes, and Ramsey]{ogarrio2016hybrid}
Juan~Miguel Ogarrio, Peter Spirtes, and Joe Ramsey.
\newblock A hybrid causal search algorithm for latent variable models.
\newblock In \emph{Conference on probabilistic graphical models}, pages 368--379. PMLR, 2016.

\bibitem[Olsson(1979)]{olsson1979maximum}
Ulf Olsson.
\newblock Maximum likelihood estimation of the polychoric correlation coefficient.
\newblock \emph{Psychometrika}, 44\penalty0 (4):\penalty0 443--460, 1979.

\bibitem[Ramsey et~al.(2017)Ramsey, Glymour, Sanchez-Romero, and Glymour]{ramsey2017million}
Joseph Ramsey, Madelyn Glymour, Ruben Sanchez-Romero, and Clark Glymour.
\newblock A million variables and more: the fast greedy equivalence search algorithm for learning high-dimensional graphical causal models, with an application to functional magnetic resonance images.
\newblock \emph{International journal of data science and analytics}, 3:\penalty0 121--129, 2017.

\bibitem[Roos(2017)]{Roos2017}
Teemu Roos.
\newblock \emph{Minimum Description Length Principle}, pages 823--827.
\newblock Springer US, Boston, MA, 2017.
\newblock ISBN 978-1-4899-7687-1.
\newblock \doi{10.1007/978-1-4899-7687-1_894}.
\newblock URL \url{https://doi.org/10.1007/978-1-4899-7687-1_894}.

\bibitem[Scutari(2014)]{scutari2014bayesian}
Marco Scutari.
\newblock Bayesian network constraint-based structure learning algorithms: Parallel and optimised implementations in the bnlearn r package.
\newblock \emph{arXiv preprint arXiv:1406.7648}, 2014.

\bibitem[Silva(2005)]{silva2005automatic}
Ricardo Silva.
\newblock \emph{Automatic discovery of latent variable models}.
\newblock Carnegie Mellon University, 2005.

\bibitem[Spirtes(1996)]{spirtes1996discovering}
Peter Spirtes.
\newblock Discovering causal relations among latent variables in directed acyclic graphical models.
\newblock 1996.

\bibitem[Spirtes et~al.(2000)Spirtes, Glymour, and Scheines]{spirtes2000causation}
Peter Spirtes, Clark~N Glymour, and Richard Scheines.
\newblock \emph{Causation, prediction, and search}.
\newblock MIT press, 2000.

\bibitem[Tsamardinos et~al.(2006)Tsamardinos, Brown, and Aliferis]{tsamardinos2006max}
Ioannis Tsamardinos, Laura~E Brown, and Constantin~F Aliferis.
\newblock The max-min hill-climbing bayesian network structure learning algorithm.
\newblock \emph{Machine learning}, 65:\penalty0 31--78, 2006.

\end{thebibliography}

\newpage
\onecolumn
\newpage
\aistatstitle{Supplementary Materials}



\section{ADDITIONAL EXPERIMENTS}
\subsection{Pairwise Log-Likelihood Correlation Estimate}\label{supp:A1}
\begin{figure}[H]
\centering
    \includegraphics[width=0.7\linewidth]{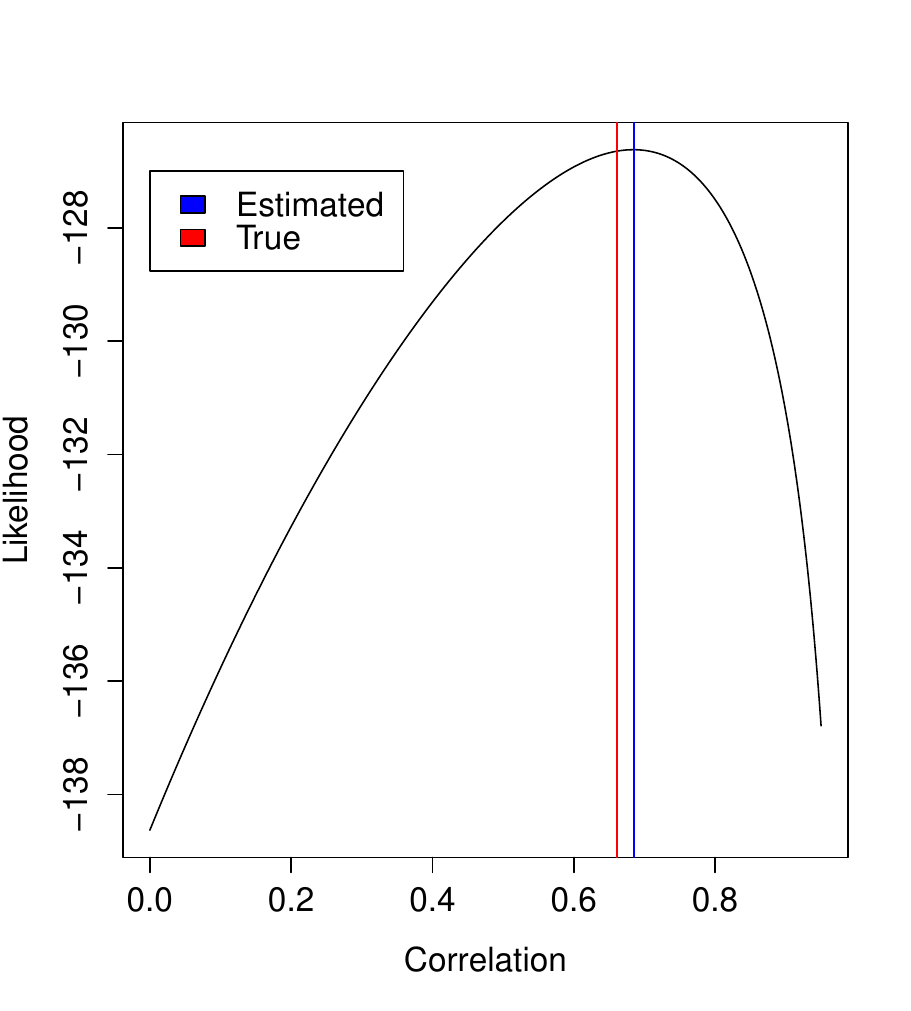}
    \caption{Single simulation of finding the correlation over a pair of units where $n = 500$ and $p = 500$.}
    \label{fig:rho_likelihood_sim}
\end{figure}

In Section~\ref{sec:cov_est}, we detail a covariance estimation method that does pairwise correlation estimates between two rows of discrete data. Figure~\ref{fig:rho_likelihood_sim} illustrates a single correlation estimation in a random DAG simulation with parameters $n = 500, p = 500$. The true correlation is indicated by the red line and the estimated maximum likelihood estimate of Equation~\ref{alg:likelihood_cov_mat} is indicated by the blue line. Our correlation estimate is very close to the true correlation.

\subsection{RMSE Covariance Estimate}\label{supp:A6}
In Section~\ref{sec:data_sim}, we assess the accuracy of the covariance matrix estimate from our pairwise estimation method. In each of the four settings, we run ten simulations and run the covariance estimation method. Given that each pairwise correlation entails a maximum likelihood estimate with $p$ samples as in~\eqref{alg:likelihood_cov_mat}, the accuracy of covariance estimates in simulations with larger $p$ values resulted in lower RMSEs.

To assess the accuracy of our covariance estimate, the root mean-squared error (RMSE) of the estimated covariance matrix $\wh \Sigma$ was calculated relative to the true covariance $\Sigma^*$:
\noindent
\begin{align*}
    \text{RMSE} (\wh \Sigma , \Sigma^*) = \left\{ \frac{1}{|H|}\sum_{(i,j)\in H} (\wh \Sigma_{ij} - \Sigma_{ij}^*)^2\right\}^{1/2},
\end{align*}
\noindent
where $H$ is the set of non-diagonal, non-zero elements of $\Sigma^*$. The RMSE is calculated among non-zero elements according to the block diagonal structure of $\Sigma^*$ mentioned in Section \ref{sec:cov_est}. Accordingly, we only estimate correlations between rows in the same cluster. Note that $\diag(\Sigma) = I$ due to the identifiability issue discussed in Section $\ref{sec:deplatentmodel}$. This metric ensures a focused evaluation of the accuracy with respect to relevant elements of the covariance matrix. Supplementary Material~\ref{supp:A6} shows the distribution of the RMSE values for various parameter sets of $n$ and $p$ where consistently lower RMSE values are observed for larger variable size $p$.

Let us consider the time complexity of this pairwise covariance estimation approach. Suppose the sample size (number of units) is $n_b$ for block $b$ and there are a total of  $B$  blocks. Then it is easy to see that the time complexity of our method is on the order of $\mathcal{O}(\sum_{b = 1}^B {n^2_b})$. For well-balanced block sizes, where $n$ is the total sample size, the time complexity is $\mathcal{O}({n^2}/{B})$.

\begin{figure}[H]
    \centering
    \includegraphics[width=.7\linewidth]{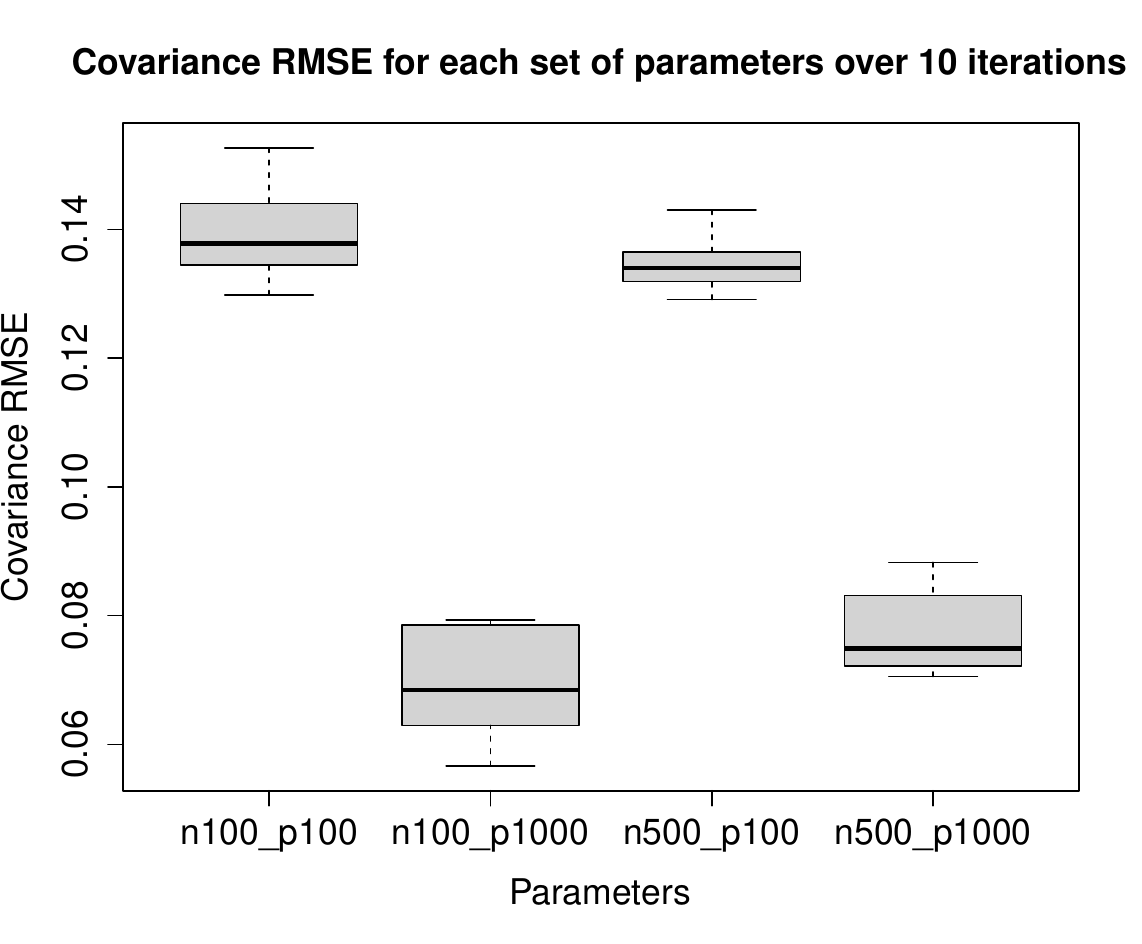}
    \caption{RMSE of estimated $\wh \Sigma$. There are 10 different simulations done corresponding to each box-plot. Simulations used a mixed covariance structure under block sizes ranging from 10 to 15 under a random DAG setting.} 
    \label{fig:cov_rmse}
\end{figure} 

\newpage

\subsection{Convergence of Algorithm~\ref{alg:discdag}}\label{supp:A2}
\begin{figure}[H]
    \centering
    \includegraphics[width=0.6\linewidth]{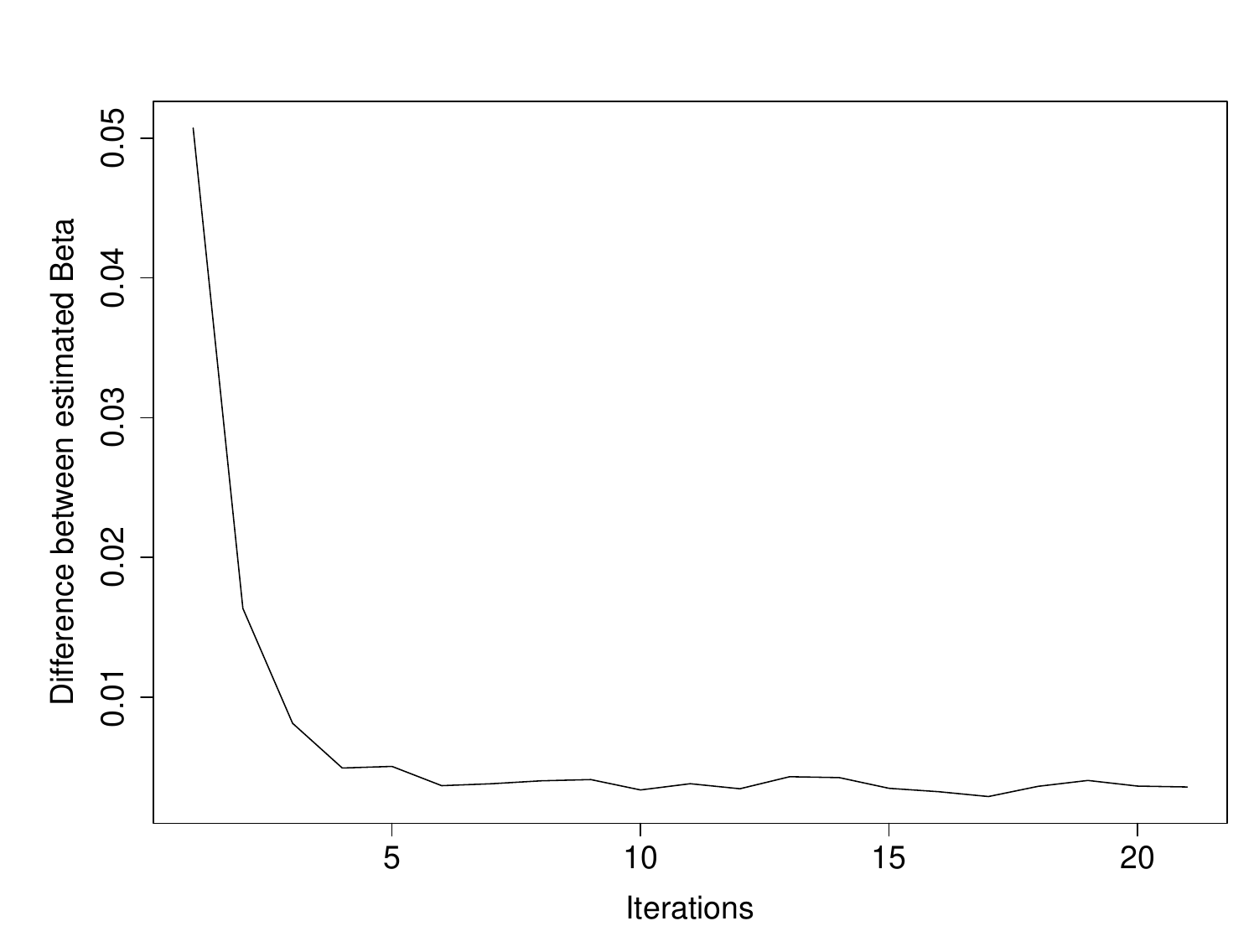}
    \caption{The difference $\lVert \beta^{(t+1)} - \beta^{(t)}\rVert$ between betas for every iteration in a simulation with $n = 100$ and $p = 100$.}
    \label{fig:converge}
\end{figure}

Section~\ref{sec:disc_decor_alg} describes the latent data recovery from discrete data and decorrelation of the dependencies among the units. We describe the iterative Algorithm~\ref{alg:discdag} that aims to obtain better estimates of $\beta$ to improve recovery of latent data and thus, more accurate estimates of the decorrelated latent data. Figure~\ref{fig:converge} shows the difference in $\beta$ between subsequent iterations is converging.
\noindent
\newpage
\subsection{Experiments on simulated data under random DAGs}\label{supp:randomDags}
For simulations, we use $n \in\{100, 500\}$ units and $p \in \{100, 1000\}$ variables. Random DAGs with $p$ nodes were fixed to $2p$ edges and edge weights were uniformly sampled from the interval $[-0.9,-0.6] \cup [0.6,0.9]$. 

\begin{figure}[H]
    \centering
    \includegraphics[width=1\linewidth]{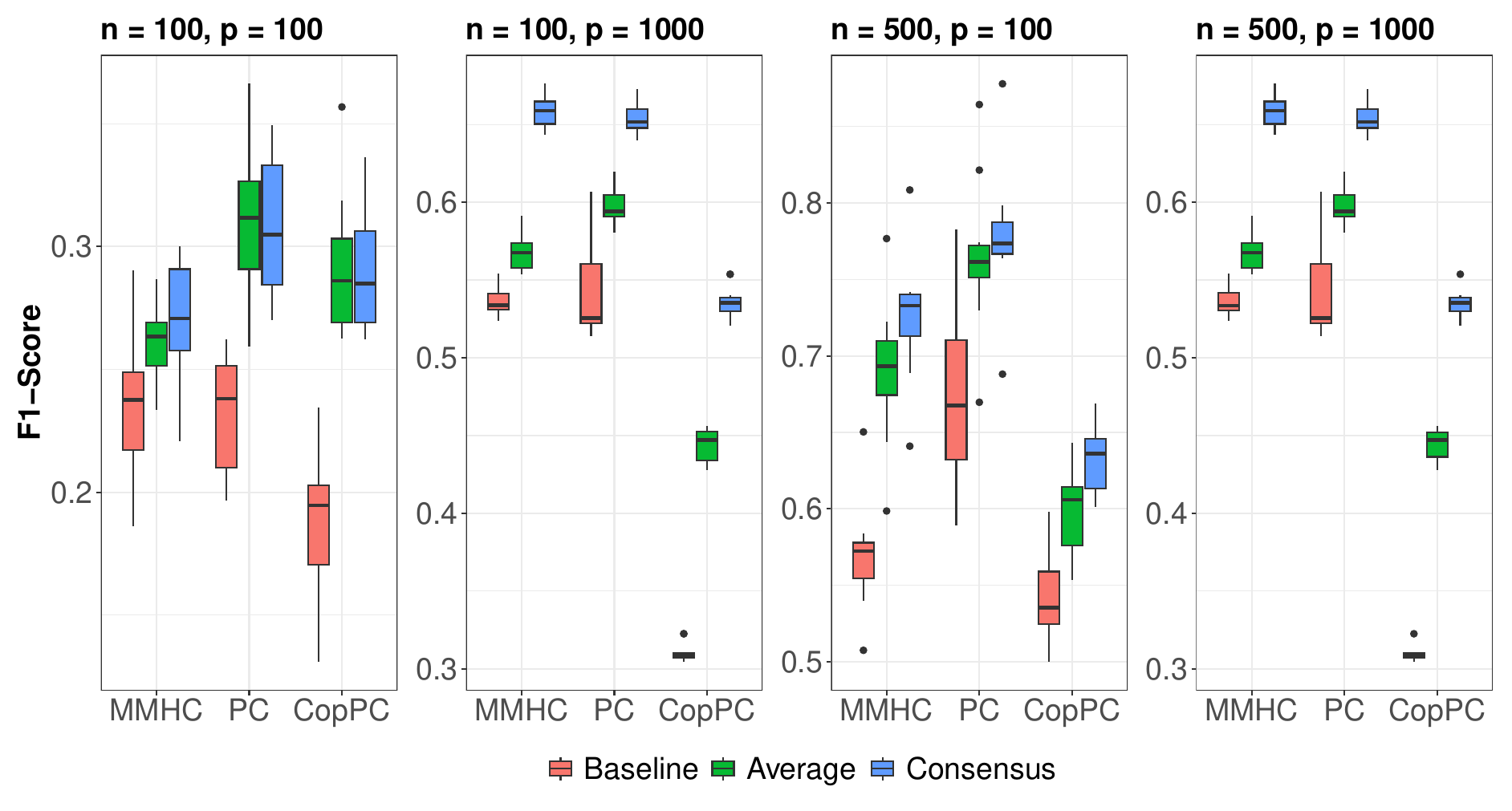}
    \caption{F-1 scores across 10 different simulations under each setting of $(n,p)$ using either MMHC, PC, or Copula-PC as the DAG learning method.}
    \label{fig:simulated_hist_f1}
\end{figure}

For visual comparison, we report the box-plots of the F-1 score for four combinations of $n$ and $p$ in Figure~\ref{fig:simulated_hist_f1}, which includes scenarios of both $p > n$ and $p < n$. For each parameter set, we conduct 10 simulations employing the MMHC method as both the initial and final DAG learning approach, and 10 simulations using the PC, and Copula-PC method in the same manner, across three approaches (Baseline, Average, Consensus).

Computation time for Algorithm~\ref{alg:discdag} and structure learning is dependent on the number of units, variables, and method. For the largest case ($n = 500, p = 1000$), using the MMHC method, a single experiment takes approximately 12 hours using an internal cluster over two nodes with 32gb of memory. For the smallest case ($n = 100, p = 100$), our method takes approximately 1 or 2 minutes under the same computational settings.

\newpage
\subsection{Covariance estimate robustness to initial DAG learning method}\label{supp:A3}

Table~\ref{RMSE-sim-table} details the RMSE for covariance estimates to the true for different covariance structures ($\Sigma$-structures) in different settings of sample and variable size $(n,p)$. The covariance estimation method relies on an initial estimate of $\beta$. Thus, we try two methods(Neighborhood Lasso and MMHC) for the initial estimate of the dependent discrete data. Neighborhood Lasso is the Lasso regression of $X_j \sim X_{-j}$ for all $j \in [p]$. Neighborhood Lasso does not estimate a DAG but rather a Markov Blanket \citep{meinshausen2006high}. Because there is no significant difference in RMSE between the two methods, we opted to use MMHC as it estimates a DAG which is consistent with our resulting output. 

After obtaining the parent estimates $\wh{PA_j}$, the initial estimate of $\beta$ for the covariance estimation method is done through Algorithm~\ref{alg:discdag} with $\wh\Sigma = I_n$. After the initial estimate of $\beta$, we then run the covariance estimation method followed by Algorithm~\ref{alg:discdag} using the estimated covariance $\wh\Sigma$.

\begin{table}[h]
\caption{RMSE of Covariance Estimate between Neighborhood Lasso and MMHC} \label{RMSE-sim-table}
\begin{center}
\begin{tabular}{cccc}
\hline 
\textbf{$\Sigma$-Structure}  &\textbf{(n,p)} &\textbf{Neighborhood Lasso RMSE} &\textbf{MMHC RMSE}\\
\hline 
\textbf{Toeplitz} & (100,100) & 0.122 & 0.130 \\ 
 & (100,100)   & 0.056 & 0.060 \\
 & (500,100) & 0.112 & 0.109 \\
 & (500,1000)   & 0.055 & 0.053 \\
\textbf{Equal} & (100,100) & 0.115 & 0.108\\
 & (100,1000)   & 0.039 & 0.039 \\
 & (500,100) & 0.108 & 0.106 \\
 & (500,1000)   & 0.041 & 0.041 \\
 \hline
\end{tabular}
\end{center}
\end{table}

\newpage
\subsection{Details on random DAG results}\label{supp:A4}
Table~\ref{sim-table} gives some additional results to Figure~\ref{fig:simulated_hist_f1}. We include three covariance structures (Mixed, Toepltiz, Equal) described in Section~\ref{sec:data_sim} with numerous sets of sample and variable size ($(n,p)$). The \textit{baseline} approach estimates the DAG based on the dependent discrete data that assumes i.i.d data. The \textit{average} and \textit{consensus} approaches use the decorrelated latent data to estimate the final DAG. The bolded numerical results indicate the best method for the set of covariance structure and $(n,p)$ and the '\% Increase' is the increase in F-1 score compared to the baseline approach. In this table, we use MMHC as the structure learning method on both the dependent binary data and continuous decorrelated data.

\begin{table}[H]
    \caption{Simulated random DAG F-1 scores of three different methods: Baseline, Average, and Consensus. The respective \% increase over the baseline for the best performing method (in bold) is given.  Each number refers to the average over 10 different simulations.}\label{sim-table}
    \begin{center}
    \begin{tabular}{cccccc}
    \hline
    \textbf{$\Sigma$-Structure}  &\textbf{(n,p)} &\textbf{Baseline} &\textbf{Average} &\textbf{Consensus} &\textbf{\% Increase}\\
     \hline
 \textbf{Mixed} & (100,100) & 0.235 & 0.261 & \textbf{0.271}  & 15.3\\
 & (100,500)   & 0.144 & 0.174 & \textbf{0.181} &  25.7 \\
 & (100,1000) & 0.116 & 0.135 & \textbf{0.143} & 23.3 \\
 & (500,100)   & 0.570 & 0.690 & \textbf{0.725} & 27.2 \\
 & (500,500) & 0.548 & 0.605 & \textbf{0.658} & 23.5 \\
 & (500,1000)   & 0.537 & 0.605 & \textbf{0.677} & 22.5 \\
\textbf{Toeplitz} & (100,100) & 0.231 & 0.148 & \textbf{0.260} & 12.6\\
 & (100,500)   & 0.144 & \textbf{0.255} & 0.175 & 21.5 \\
 & (100,1000) & 0.115 & \textbf{0.160} & 0.140 & 21.7 \\
 & (500,100)   & 0.630 & 0.664 & \textbf{0.727} & 25.1 \\
 & (500,500) & 0.580 & 0.619 & \textbf{0.696} & 22.1 \\
 & (500,1000)   & 0.529 & 0.569 & \textbf{0.602} & 20.6\\
\textbf{Equal} & (100,100) & 0.168 & 0.278 & \textbf{0.283} & 28.6\\
 & (100,500)   & 0.139 & 0.167 & \textbf{0.182} & 33.8 \\
 & (100,1000) & 0.086 & 0.144 & \textbf{0.153} & 33\\
 & (500,100)   & 0.449 & 0.660 & \textbf{0.712} & 24.9 \\
 & (500,500) & 0.372 & 0.550 & \textbf{0.613} & 15 \\
 & (500,1000)  & 0.327 & 0.564 & \textbf{0.649} & 25.3 \\
 \hline
    \end{tabular}
    \end{center}
\end{table}

\vfill

\end{document}